\documentclass[letterpaper,times]{IONconf}

\usepackage[pdftex]{graphicx}

\usepackage[justification=centering]{caption}
\graphicspath{{pictures/}}
\DeclareGraphicsExtensions{.pdf,.jpeg,.png}
\usepackage{array}
\usepackage{titlesec}
\usepackage{titlesec}
\usepackage{amsmath}
\usepackage{amssymb}
\usepackage{algorithm,algorithmicx,algpseudocode,amsmath,bm}
\usepackage{siunitx}

\usepackage{multirow}
 \usepackage[table,xcdraw]{xcolor}

\DeclareMathOperator{\ATE}{ATE}
\DeclareMathOperator{\RPE}{RPE}
\algnewcommand\INPUT{\item[\textbf{Input:}]}
\algnewcommand\OUTPUT{\item[\textbf{Output:}]}
\newcolumntype{P}[1]{>{\centering\arraybackslash}p{#1}}

\usepackage{footnote}
\makesavenoteenv{table} 
\makesavenoteenv{tabular}
\usepackage{array}
\usepackage{url}

\usepackage[%
backend=biber,
natbib=true,
style=apa,
]{biblatex}
\usepackage{amsmath} 
\usepackage[hidelinks]{hyperref}
\usepackage{siunitx}
\usepackage{booktabs}
\usepackage{gensymb}
\usepackage[normalem]{ulem}
\usepackage{color}
\usepackage{colortbl}
\usepackage[super]{nth}

\newcommand{\Reals}{\mathbb{R}}

\title{A Robust Localization Solution for an Uncrewed Ground Vehicle in Unstructured Outdoor GNSS-Denied Environments}

\author{
    W. Jacob Wagner, Isaac Blankenau, Maribel DeLaTorre, Amartya Purushottam,
    and Ahmet Soylemezoglu%
    \vspace{1mm} \\%
    \textit{Engineer Research and Development Center (ERDC)}% <- this '%' removes a trailing whitespace
    }

\begin{document}

\maketitle

\begin{figure}[h]
    \centering
    \includegraphics[width=5in]{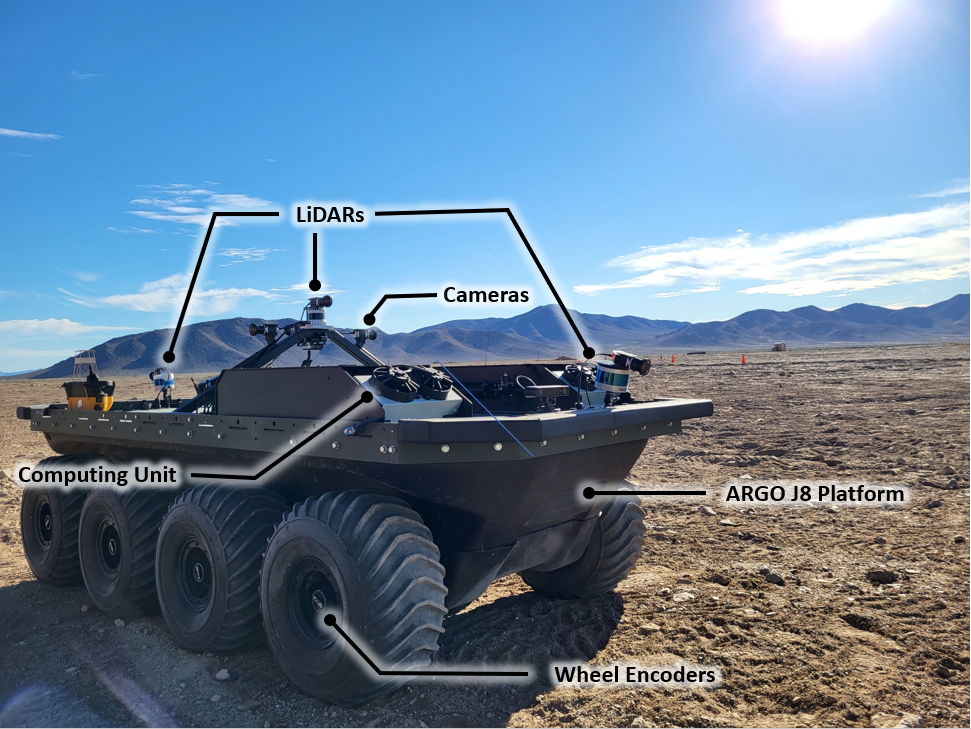}
    \caption{The Robotics for Engineer Operations (REO) Speedwagon site characterization UGV platform. IMU's housed inside chassis.}
    \label{fig:j8}
\end{figure}

\section*{Biographies}
\textbf{W. Jacob Wagner} is a Research Engineer at ERDC working on autonomy and robotics problems including localization in GNSS-denied environments and development of learning-based control systems for autonomous earthmoving equipment. He is also a Ph.D. candidate in the Electrical and Computer Engineering (ECE) program at the University of Illinois at Urbana-Champaign (UIUC) and a member of the Human Centered Autonomy Laboratory.

\textbf{Isaac Blankenau} is a Research Mechanical Engineer at ERDC and a technical manager on the Robotics for Engineer Operations (REO) program. He works on autonomy and robotics problems, focusing on localization in GNSS-denied environments.

\textbf{Maribel DeLaTorre} is a Research Mechanical Engineer at ERDC. She works on autonomy and sensor integration for localization systems, including systems meant for GNSS-denied environments. 

\textbf{Amartya Purushottam} is a Research Engineer at ERDC working on estimation and control problems in robotics including filtering for GNSS-denied localization. He is also a Ph.D. candidate in the ECE program at UIUC working on shared teleoperation of humanoid robotics at the RoboDesign Lab.

\textbf {Ahmet Soylemezoglu, Ph.D.}is a Research Systems Engineer at ERDC and currently serves as the program manager for the Robotics for Engineer Operations (REO) research and development program. His primary research interests are model based approaches to complex-adaptive-systems design and analysis; model based systems engineering; distributed and decentralized control of autonomous systems; and machine learning.

\section*{Abstract}
This work addresses the challenge of developing a localization system for an uncrewed ground vehicle (UGV) operating autonomously in unstructured outdoor Global Navigation Satellite System (GNSS)-denied environments. The goal is to enable accurate mapping and long-range navigation with practical applications in domains such as autonomous construction, military engineering missions, and exploration of non-Earth planets. The proposed system - Terrain-Referenced Assured Engineer Localization System (TRAELS) – integrates pose estimates produced by two complementary terrain referenced navigation (TRN) methods with wheel odometry and inertial measurement unit (IMU) measurements using an Extended Kalman Filter (EKF). Unlike simultaneous localization and mapping (SLAM) systems that require loop closures, the described approach maintains accuracy over long distances and one-way missions without the need to revisit previous positions. Evaluation of TRAELS is performed across a range of environments. 
In regions where a combination of distinctive geometric and ground surface features are present, the developed TRN methods are leveraged by TRAELS to consistently achieve an absolute trajectory error of less than \SI{3.0}{\meter}. % 
The approach is also shown to be capable of recovering from large accumulated drift when traversing feature-sparse areas, which is essential in ensuring robust performance of the system across a wide variety of challenging GNSS-denied environments.
Overall, the effectiveness of the system in providing precise localization and mapping capabilities in challenging GNSS-denied environments is demonstrated and an analysis is performed leading to insights for improving TRN approaches for UGVs.

\section{Introduction}
In applications such as autonomous construction and site characterization \parencite{Soylemezoglu2021}, UGVs are viable platforms for interacting with the environment and assessing terrain traversability.  For example, while traveling through unexplored regions, UGVs can be equipped with sensors to evaluate soil conditions. To travel long distances autonomously, they must be capable of high fidelity global navigation and accurate mapping. As such, these vehicles typically house onboard localization solutions that rely on a combination of techniques such as dead-reckoning (DR), GNSS corrections, SLAM \parencite{SLAM_he2020integrated}, and TRN \parencite{TRN_review}. While the particular choice and combination of methods is often driven by application needs, many outdoor systems rely heavily on GNSS support. Basic GNSS enables global positioning within $\sim \SI{3}{\meter}$ and can achieve an order of magnitude more accurate measurements with use of real-time kinematic (RTK) positioning enabling creation of high-quality 3D maps. However, geographical hindrances such as tall buildings in urban environments or dense vegetation in rural areas, can lead to signal attenuation and multipath errors resulting in positioning  inaccuracies. Additionally, GNSS availability is susceptible to intentional disruptions such as jamming or spoofing. 

In the absence of GNSS, either DR or SLAM algorithms have been widely utilized to enable autonomous operations. However, both fundamentally result in the accumulation of positioning errors. State-of-the-art SLAM systems generate maps that are locally referenced and exhibit an absolute trajectory error drift of $0.15\%$ to $0.35\%$ in unstructured environments where GNSS signals are unavailable (Meyer, 2021). They rely on loop closures during the return journey to correct the vehicle's pose and refine the map estimate. However, prior to the loop-closure corrections (i.e. when traveling directly to a distant destination), real-time estimates of the robot's pose may be less accurate.

Terrain referenced navigation (TRN), also known as terrain aided navigation, can replace GNSS in challenging environments, providing absolute pose estimates that can compensate for drift. 
These techniques compare \textit{a priori} elevation maps with vehicle ranging measurements, which are both typically obtained from a top down perspective.
TRN is mainly used for aerial and submersible vehicles, where large scale terrain features are visible and distinctive \parencite{trnaerial_steerable_carroll2021terrain, Salavasidis2019}.
For UGV applications, however, the ground-based measurements are largely limited by line of sight, reducing the effectiveness of TRN methods in environments with gradual elevation changes.
Furthermore, the perspective difference between the aerially obtained \textit{a priori} map data and the ground-based measurements from the vehicle makes the matching process more challenging.
Therefore, ground platforms often resort to ground-to-aerial view matching algorithms, which compare vehicle obtained imagery with aerial or satellite imagery.
For example, \cite{Cheung2018} use ground-level panoramas and satellite imagery for long-range UGV localization, but only estimate the vehicle’s position, not the complete pose. Similarly, \cite{Hu2020} use a convolutional neural network to match ground-view and satellite images for geo-referencing, but only in urban areas with roads and prominent features. They do not address unstructured environments.
In this paper, a broader definition of terrain that includes not only relief, but also surface characteristics such as color and land cover is used. The term TRN is therefore used to encompass all techniques that leverage \textit{a priori} maps of large outdoor areas to enhance localization.

Overall, the availability of GPS under environmental constraints (terrain, tree cover, GPS spoofing) and the accuracy of DR, SLAM and TRN over long distances for UGV applications are not consistently guaranteed. Accumulation of error eventually leads to inaccurate navigation and hinders the successful exploration of the environment. This work addresses the challenge of developing a localization system for a UGV autonomously navigating over long ranges and performing mapping in unstructured outdoor Global Navigation Satellite System \hbox{(GNSS)-denied} environments.
Specifically, a localization framework is introduced that leverages two asynchronous extended Kalman filters (EKFs) for combining TRN state estimates, with odometry and IMU measurements.

The main contributions of this work are:
\begin{itemize}
    \item Development of robust real-time localization framework, Terrain-Referenced Assured Engineer Localization System (TRAELS), for UGV applications in unstructured environments using: (i) two TRN methods, (ii) a wheel slip detection and rejection module, (iii) IMU and gyro-compassing INS measurements, (iv) a yaw bias estimation filter, (v) a sensor calibration procedure to minimize dead-reckoning error, and (vi) extended Kalman filters to fuse the measurements for global and local navigation.   
    \item Validation of TRAELS across four diverse GNSS-denied environments, where an absolute trajectory error of less than \SI{3.0}{\meter} is consistently achieved for conducive environments
    \item Analysis of the system performance highlighting scenarios where the individual TRN methods excel, and under what conditions they are not sufficient to maintain good performance. 
\end{itemize}

\section{Methods}\label{sec:methods}

\begin{figure}[t!]
    \centering
    \includegraphics[width=7in]{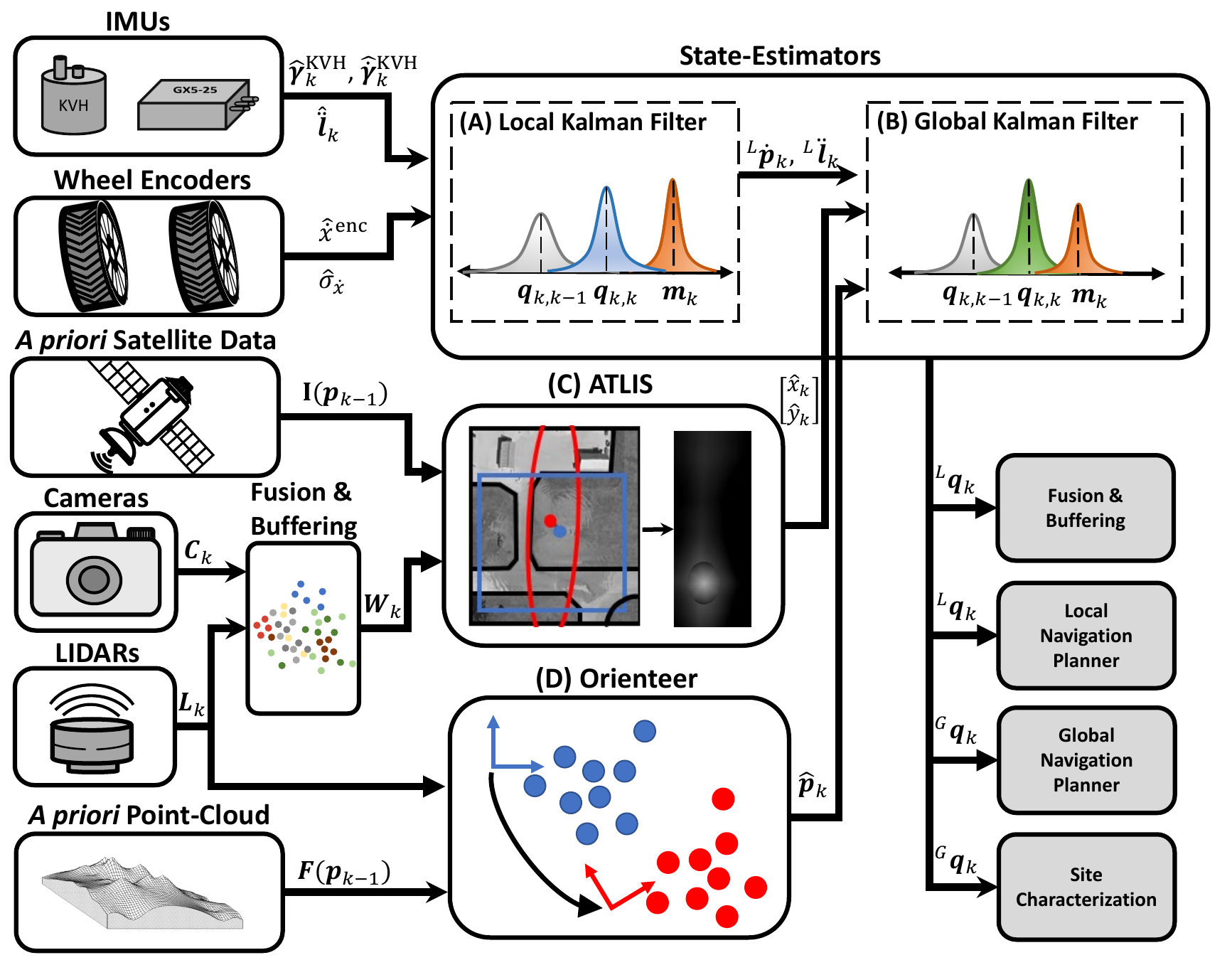}
    \caption{Localization pipeline and system overview. (A) outputs a complete state estimate - ${}^{L}\boldsymbol{q}_k$ - in the local frame where no global corrections are provided ensuring smoothness of the state required by the local navigation planner and fusion \& buffering system. (B) fuses together global position measurements from terrain-referenced pose estimation methods to produce ${}^{G}\boldsymbol{q}_k$, which is used for global navigation and site characterization but is susceptible to larger discontinuities. ATLIS and Orienteer provide global pose estimates by aligning locally obtained point-clouds with (C) \textit{a priori} satellite imagery using template matching and (D) point-clouds using registration techniques.}
    \label{fig:block_diag}
\end{figure}

The goal of the system is to enable a UGV to perform accurate voxel-based mapping and long-range autonomous navigation in unstructured outdoor environments without the use of GNSS. 
The approach described ensures optimal dead-reckoning performance through the use of low-noise IMUs, accurate calibration of sensor covariances and mounting orientations, and the implementation of a wheel slip rejection method.
The accumulation of drift is addressed by two TRN methods, Active Terrain Localization Imaging System (ATLIS) and Orienteer, within our global estimator as shown in Fig \ref{fig:block_diag} and aided by online estimation of yaw bias.

Mapping is performed using a voxel-based approach which enables the compression of point-clouds, seamless integration with \textit{a priori} data, and easy generation of standard geospatial output products \parencite{Richards2022}.
Due to the lossy nature of this map representation, it is difficult to integrate with SLAM methods which can update the map backward in time using bundle adjustments.
Therefore, an online TRN approach provides localization using an EKF, although components of the system could also be used to improve the mapping performance of a SLAM system.

\subsection{Platform Overview} \label{sec:overview}
The all-terrain UGV platform used in this investigation is an ARGO J8 ATLAS Xtream Terrain Robot (XTR) that has been substantially modified and equipped with various sensors, including LIDARs, cameras, and inertial measurement unit (IMU) sensors, see Figure \ref{fig:j8} and Table \ref{tab:j8_hardware}.
The software system is built on the Robot Operating System (ROS) Noetic, comprised of approximately 100 ROS packages, both open source and internally developed.
\begin{table}
    \centering
    \def\arraystretch{1.5}
        \begin{tabular}{|ccc|}
            \hline
            \rowcolor[HTML]{9B9B9B} 
            \multicolumn{3}{|c|}{\cellcolor[HTML]{9B9B9B}\textbf{Hardware}} \\ \hline
            \rowcolor[HTML]{C0C0C0} 
            \multicolumn{1}{|c|}{\cellcolor[HTML]{C0C0C0}} & \multicolumn{1}{c|}{\cellcolor[HTML]{C0C0C0}\textbf{Vehicle}} & ARGO Atlas J8 XTR UGV \\ \cline{2-3} 
            \rowcolor[HTML]{C0C0C0} 
            \multicolumn{1}{|c|}{\multirow{-2}{*}{\cellcolor[HTML]{C0C0C0}\textbf{Platform}}} & \multicolumn{1}{c|}{\cellcolor[HTML]{C0C0C0}\textbf{Wheel Odometry}} & Magnetic Hall Effect \\ \hline
            \rowcolor[HTML]{FFCCC9} 
            \multicolumn{1}{|c|}{\cellcolor[HTML]{FFCCC9}} & \multicolumn{1}{c|}{\cellcolor[HTML]{FFCCC9}\textbf{IMU}} & 2x Parker LORD MicroStrain 3DMGX5-AHRS \\ \cline{2-3} 
            \rowcolor[HTML]{FFCCC9} 
            \multicolumn{1}{|c|}{\multirow{-2}{*}{\cellcolor[HTML]{FFCCC9}\textbf{Inertial}}} & \multicolumn{1}{c|}{\cellcolor[HTML]{FFCCC9}\textbf{INS}} & 1x KVH GEO-FOG 3D \\ \hline
            \rowcolor[HTML]{DAE8FC} 
            \multicolumn{1}{|c|}{\cellcolor[HTML]{DAE8FC}} & \multicolumn{1}{c|}{\cellcolor[HTML]{DAE8FC}\textbf{LiDAR}} & 1x Ouster OS-64 \\ \cline{2-3} 
            \rowcolor[HTML]{DAE8FC} 
            \multicolumn{1}{|c|}{\cellcolor[HTML]{DAE8FC}} & \multicolumn{1}{c|}{\cellcolor[HTML]{DAE8FC}\textbf{LiDAR}} & 2x Velodyne Pucks VLP-16 \\ \cline{2-3} 
            \rowcolor[HTML]{DAE8FC} 
            \multicolumn{1}{|c|}{\multirow{-3}{*}{\cellcolor[HTML]{DAE8FC}\textbf{Perception}}} & \multicolumn{1}{c|}{\cellcolor[HTML]{DAE8FC}\textbf{Camera}} & 6x Basler ace a2A1920-52gcBAS \\ \hline
            \rowcolor[HTML]{9AFF99} 
            \multicolumn{1}{|c|}{\cellcolor[HTML]{9AFF99}\textbf{Ground-Truth}} & \multicolumn{1}{c|}{\cellcolor[HTML]{9AFF99}\textbf{RTK Navigation Solution}} & \begin{tabular}[c]{@{}c@{}}1x Parker LORD MicroStrain 3DM-GQ7-GNSS/INS \\ SensorCloud\footnote{https://rtk.sensorcloud.com/} RTK corrections \end{tabular} \\ \hline
            \rowcolor[HTML]{FFFFC7} 
            \multicolumn{1}{|c|}{\cellcolor[HTML]{FFFFC7}\textbf{Computing}} & \multicolumn{1}{c|}{\cellcolor[HTML]{FFFFC7}\textbf{Computer Specifications}} & \begin{tabular}[c]{@{}c@{}}4x Computers:\\ Intel Core i9 Extreme Edition Processor\\ WS X299 Sage/10G Motherboard\\ 128 GB RAM\\ OS: Ubuntu 20.04\end{tabular} \\ \hline
        \end{tabular}
    \caption{Key hardware used on the site characterization platform REO-Speedwagon}
    \label{tab:j8_hardware}
\end{table}

\subsection{A Priori Data} \label{sec:aPriori}
The TRN methods produce global corrections by aligning locally obtained data to \textit{a priori} maps.
In the experiments performed, this data is collected from a variety of sources, derived from different sensing systems, collected on different dates, and times of year and day, resulting in maps of varying quality and accuracy.
In maps downloaded directly from USGS database, inaccuracies of over 1 meter have been observed which lead to higher TRN estimation errors.
In the case that color imagery is obtained separately from spatial data, offsets between the data can additionally lead the normally complementary TRN methods to disagree, producing oscillating positioning.
To overcome these issues, all map products are manually aligned to Google Earth.
While this is unlikely to improve the overall accuracy of the maps, it does increase agreement between the two TRN methods. 

Raster elevation data, such as digital elevation models (DEMs), cannot represent vertical surfaces like building walls or tree trunks. Additionally, the density of point-clouds obtained from aerial platforms is low on these surfaces. However, a UGV’s LiDAR sensors produce point clouds that mainly consist of ground and vertical surfaces with no points on the top of structures. This perspective difference makes lateral localization using the Orienteer TRN method (see section \ref{sec:orienteer}) challenging. To address this issue, the \textit{a priori} spatial data is meshed and re-sampled to ensure a sufficient minimum density on all vertical surfaces. This process can create artifacts, such as sloped walls or shifts in vertical surfaces due to overhangs, but improves matching performance significantly.

\subsection{State Estimation} \label{sec:onlineStateEst}
To enable global state estimation, a relative inertial frame is defined at the starting position of the vehicle, referred to as the global frame $G$.
Using an initial global pose (easting, northing, and absolute heading), a static transform, $T_G^{UTM}$, between the nearest Universal Transverse Mercator (UTM) grid and global frame ties the inertial frame to a global coordinate reference system.
The use of this transform enables globally referenced mapping and specification of waypoints as easting and northing values, which are needed to allow users to create routes with respect to a satellite map.

The EKF estimates the UGV state in generalized coordinates $\boldsymbol{q} = [\boldsymbol{p} \; \dot{\boldsymbol{p}} \; \ddot{\boldsymbol{l}}]^\top \in \Reals^{15}$ where $\boldsymbol{p} = [\boldsymbol{l}, \boldsymbol{\gamma}]^\top \in \Reals^6$ represents the vehicle's pose in the inertial frame with position $\boldsymbol{l} = [x, y, z]^\top$ and orientation $\boldsymbol{\gamma} = [\phi, \theta, \psi]^\top$.
The twist $\dot{\boldsymbol{p}} \in \Reals^6$ is tracked in vehicle frame $V$ as is the linear acceleration $\ddot{\boldsymbol{l}} \in \Reals^3$.
The EKF uses an omnidirectional kinematic motion model \parencite{Moore2016} where direct measurements of a subset of the system state $\mathbf{q}$ is assumed.
The filter enables the fusion of asynchronous sensor measurements to produce full state estimates.

In order to navigate between waypoints locally and smoothly track planned trajectories, a continuous frame is essential. However, considering the requirements for accurate long-distance navigation, necessary position corrections from TRN methods may cause discontinuous jumps in the desired trajectory.
To address these competing demands, we employ two asynchronous state estimators—one that estimates the state of the robot in a local inertial frame $L$, and another that does so in global frame $G$, as depicted in Figure \ref{fig:block_diag}.

The local EKF consumes measurements of linear acceleration $\hat{\Ddot{\boldsymbol{l}}}$ from all 4 IMUs, angular velocity $\hat{\dot{\boldsymbol{\gamma}}}^\text{KVH}$ and angular position $\hat{\boldsymbol{\gamma}}^\text{KVH}$ from the GeoFOG, and vehicle $x$ velocity $\dot{x}^\text{enc}$ from wheel odometry to produce a complete state estimate ${}^{L}\boldsymbol{q}_k$ at 100\unit{\hertz} that is continuous in pose and twist.
To better capture the dynamics of the skid steer platform without modifying the process model, zero-value measurements of $\dot{y}$ and $\dot{z}$ are additionally provided to the EKF.

The twist $\dot{\boldsymbol{p}}$ and linear acceleration $\dot{\boldsymbol{l}}$ estimate from the local EKF is provided to the global EKF as a measurement, which additionally fuses in ATLIS $\hat{\boldsymbol{l}}_{xy}^\text{ATLIS}$ and Orienteer $\hat{\boldsymbol{p}}^\text{Orienteer}$ pose estimates to produce ${}^{G}\boldsymbol{q}$ at a lower frequency of 10\unit{\hertz}.
As the UGV traverses the environment, the frames $L$ and $G$ diverge from one another as the result of drift from dead-reckoning in the local estimator.
The transform between these two inertial frames is maintained over time to enable transforming of measurements between frames, e.g. transforming the buffered colored point-cloud used by ATLIS from $L$ to $G$ prior to rasterization and matching.

\subsection{Calibration} \label{sec:calibration}
IMU mounting inaccuracies introduce errors in the estimation of filtered orientation, and deteriorate the performance of dead-reckoning methods. Similarly, precise determination of the wheel diameter plays a critical role in achieving accurate odometry. To address these issues, procedures were developed to accurately assess the sensor mounting orientation with respect to the vehicle's base frame and to precisely determine wheel diameter. These procedures are easy to perform and improve DR performance.

Furthermore, while commercial off-the-shelf (COTS) IMU sensors are often equipped with state estimation filters and are intended for easy installation on a variety of applications, experimentation reveals state estimates produced in challenging applications such as a skid-steer traversing rough terrain still require calibration and adjustment of their covariances due to sensor bias, imprecise mounting, and noisy vehicle dynamics.
Calibration is performed by affixing sensor to vehicle and bringing the UGV to a near-constant velocity on a smooth surface.
If the measured angular velocity, $\dot{\gamma}$, and linear acceleration , $\ddot{l}$, is off-centered, it is adjusted with an offset.
Once adjusted, the angular velocity covariance, $\sigma_{\dot{\gamma}}$, is computed as shown in (\ref{eq:dynCov}) where $\dot{\gamma} _{k}$ is the variance of the collected data.
Linear acceleration covariances, $\sigma_{\ddot{l} }$, are obtained in this manner as well. 
Using this method to estimate sensor covariances, where the vehicle is moving as opposed to stationary, helps to account for minor biases, sampling frequency errors, distortion of vehicle state measurement introduced by sensor mounting dynamics, and other challenging to characterize sources of error.
\begin{align}
    \quad
    \sigma_{\dot{\gamma} } =\frac{\sum^{N}( \dot{\gamma} _{k})^{2}}{N -1}
    \label{eq:dynCov}
\end{align}

\subsection{Heading Determination and Yaw Bias Estimation (YBE)}
Accurate attitude estimation, particularly with regard to heading, is crucial for the successful long-distance navigation of UGVs.
This becomes especially important in GPS-denied environments where estimation and control rely on dead reckoning strategies for extended periods.
For example, assuming zero-drift dead-reckoning, at a \SI{1}{\kilo\meter} distance, an error of only \SI{1}{\degree} in starting orientation results in a large positional error of \SI{17.5}{\meter}.
Most commercially available magnetometers are only able to achieve an heading accuracy on the order of $\pm$\SI{2}{\degree}, but on electric UGVs even this is not achievable due to significant time-variant magnetic interference from the vehicle's motors limiting sensor mounting locations \citep{GebreEgziabher2001AN}.

Alternatively, with the use of highly sensitive gyroscopes such as fiber optic gyroscopes (FOG) or hemispherical resonator gyroscopes (HRG), gyro-compassing based methods that use measurement of the Earth's rotation to determine heading can be used to obtain more accurate yaw estimates in magnetically challenging environments \citep{spielvogel2020adaptive}.
The GEO-FOG device used in these experiments employs a gyro-compassing method to enable estimation of the vehicle heading.
A \SI{15}{\minute} calibration procedure referred to as North-seeking is necessary to obtain the gyroscope bias estimates and a heading estimate.
In practice, the North-seeking has only attained consistent accuracy on the order of $\pm$\SI{3}{\degree} with this device, an order of magnitude lower than the specified {$\pm 0.3 \sec(\text{latitude})$}\unit{\degree} \citep{kvh_datasheet}.

To address these challenges in obtaining accurate heading, a yaw bias estimation (YBE) system was developed.
A simple additive yaw bias measurement model is assumed: $\bar{\psi} = \hat{\psi} + \Delta \psi$ where $\hat{\psi}$ is the measured UGV yaw provided by the INS and $\Delta \psi$ is the yaw bias.
Define the vector $\boldsymbol{d}_{k} = \boldsymbol{p}_{t_k} - \boldsymbol{p}_{t_{k-1}}$ as the difference in 2D position at times $t_k$ and $t_{k-1}$.
Points along each trajectory are selected such that $||{}^{G}\boldsymbol{d}_k||>\SI{50}{\meter}$ to ensure a sufficiently large signal to noise ratio of the estimate.
An additional check is performed such that only poses for which the covariance of the state estimate $\boldsymbol{p}_k$ has recently declined are included.
This helps to ensure estimation of $\Delta \psi$ is only performed when TRN methods are successfully reducing the positional uncertainty.
Fundamentally, the YBE attributes differences in the vectors ${}^{G}\boldsymbol{d}$ and ${}^{L}\boldsymbol{d}$, produced by the global and local estimators respectively, to errors in estimation of the true yaw $\bar{\psi}$ due to $\Delta \psi$.
It is important that dead-reckoning is accurate for this assumption to hold, see section \ref{sec:calibration}.
An estimate is derived by computing the angle between these vectors
\begin{equation}
    \Delta \psi_k = \arccos \left( \frac{{}^{G}\boldsymbol{d}_k \cdot {}^{L}\boldsymbol{d}_k}{||{}^{G}\boldsymbol{d}_k|| ||{}^{L}\boldsymbol{d}_k||} \right)
\end{equation}
Individual measurements of the bias $\Delta \psi_k$ are accumulated over time and a modified moving-average filter is employed to estimate the bias to account for possible time-varying biases.
This estimated bias $\Delta \hat{\psi}$ is then added to the measured heading produced by the GEO-FOG prior to consumption by the filters.

\subsection{Wheel Odometry Measurement \& Slip Rejection} \label{sec:wheelslip}
In environments with significant traction and when UGV acceleration is limited, slip free contact between UGV wheels and the ground can be ensured.
However, when UGVs navigate in unstructured environments, they encounter unpredictable ground conditions that can lead to wheel slip. If this slip is not effectively managed, it can result in rapid positional drift. To mitigate this issue, off-road vehicles often incorporate optical speed sensors \parencite{Benznavi23} which are accurate when traveling above some minimum speed.
For the mapping mission the REO Speedwagon platform is utilized for, lower speeds are necessary and optical sensors are not applicable.

Therefore, a slip rejection approach is developed to effectively increase the uncertainty in the wheel odometry measurements when the measurements disagree significantly with the current state estimate.
The covariance is increased in a step-wise fashion, as the Malhalanobis distance between the incoming odometry measurement and the current state estimate rises above a series of fixed thresholds.
This inflation is limited to ensure that accelerometer measurements can not cause the estimated velocity to diverge too much from the wheel odometry measurements ensuring system stability and allowing recovery from prolonged wheel slip.
The effectiveness of this approach is further enhanced by employing multiple high-accuracy IMUs. 

Beyond wheel slip, the accuracy of the encoder estimation of $\dot{\boldsymbol{p}}$ varies dynamically. When the vehicle is static, the encoders produce an accurate, low noise measurement of $\dot{\boldsymbol{p}}$. Conversely, in cases of high wheel load, such as rapid accelerations, the measurement accuracy is degraded. To capture this effect, wheel velocity covariances are inflated, $\hat{\sigma}_{\dot{x}}$, at higher velocities and accelerations using 
\begin{align} \label{eq:wheel_cov}
    \hat{\sigma}_{\dot{x}} =\sqrt{a+b\hat{\dot{x}} \ +\ c\hat{\ddot{x}}}
\end{align}
where $a$, $b$, and $c$ are tuning constants.
The described approach has demonstrated the ability to enhance localization accuracy in high-slip environments while imposing a negligible impact when wheel slip is insignificant.

\subsection{Terrain-Referenced Pose Estimation}
Two complementary terrain-referenced pose estimation methods are leveraged to enable robust global positioning across a variety of environments.
For both approaches, it is essential to automatically estimate the confidence in the quality of the estimated pose to enable the EKF to fuse these measurements into the state estimate properly.

\subsubsection{Active Terrain Localization Imaging System (ATLIS)}\label{sec:atlis}
\begin{figure}[ht]
    \centering
    \includegraphics[width=4in]{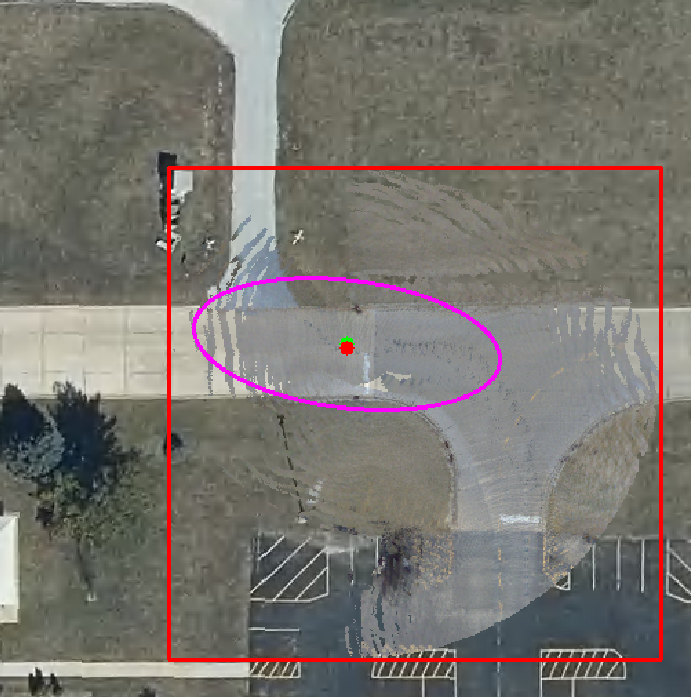}
    \caption{An example of ATLIS aligning to satellite imagery, where the estimated 2$\sigma$ covariance is shown with the pink ellipse. The red dot depicts the platform's position, and the green dot (occluded) is the position estimation from ATLIS based on the template matching approach.}
    \label{fig:atlis}
\end{figure}

ATLIS leverages template matching to estimate a vehicle's global position on an \emph{a priori} overhead image \parencite{Niles2022}.
A subset of the full \emph{a priori} is used for matching, where the patch is centered on the current position estimate and the size is varied dynamically in proportion the current estimated positional covariance $\boldsymbol{I}(\boldsymbol{p}_k)$.
The effect is that as the system becomes less certain in its position, the search region is expanded to allow for larger possible corrections.
The template image is generated by projecting a buffered colorized point-cloud $\boldsymbol{W}_k$ \parencite{Ellison2023} onto a 2D ground plane.
Environmental factors have a strong impact on localizability for terrain-referenced methods \parencite{https://doi.org/10.1002/navi.306}. In environments with distinctive visual features and high contrast, ATLIS pose estimates are generally accurate and reliable.
However, as visual contrast degrades, such as in the Desert environment shown in Figure \ref{fig:j8}, the estimates become less accurate and noisier.
ATLIS represents this uncertainty using a Gaussian distribution where the covariance is obtained by performing a weighted covariance calculation on the output of template matching based on normalized cross-correlation, as shown by the pink ellipse in Figure \ref{fig:atlis}.

\begin{figure}[ht!]
    \centering
    \includegraphics[width=7in]{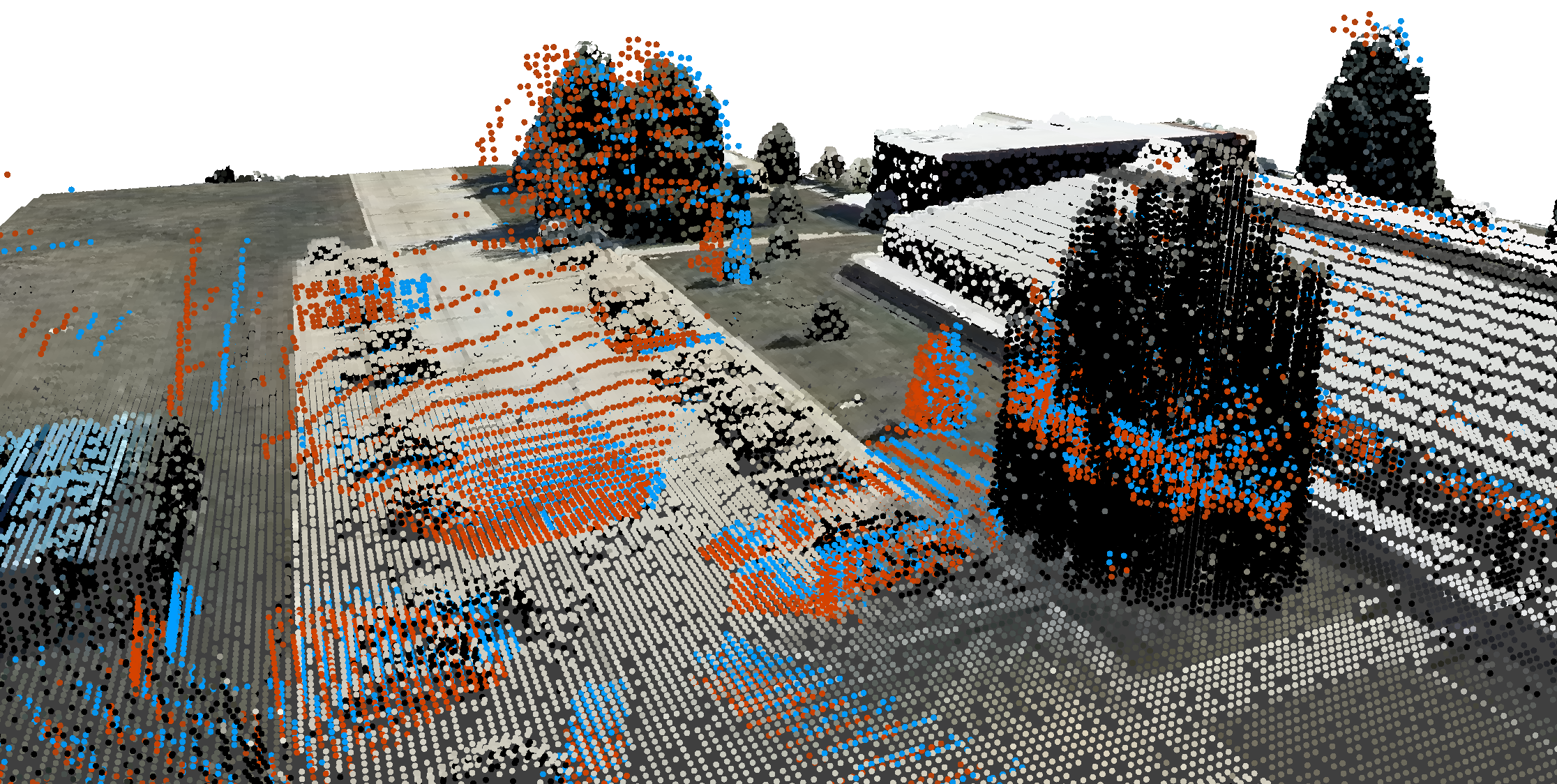}
    \caption{An example of Orienteer registering a locally obtained point-cloud (red) with the \emph{a priori} point-cloud (colored voxel map) to produce a corrected cloud (blue) at the Urban environment.}
    \label{fig:orienteer}
\end{figure}

\subsubsection{Orienteer}\label{sec:orienteer}
While ATLIS takes advantage of variations in surface color to provide an estimated global 2D position, the Site Model Geospatial System (SMGS) Orienteer system estimates a full 6D global pose by aligning point-clouds obtained from onboard sensors to \textit{a priori} georeferenced point-clouds \parencite{Richards2022}. The SMGS database is populated with \textit{a priori} geospatial information prior to executing a mission by importing DEMs, DSMs, or point-clouds of the area of interest compressing and standardizing the data into a fixed resolution voxel-based representation.

The Orienteer architecture allows for interchangeable point-cloud registration techniques, including Iterative Closest Point (ICP) and 3D Normal Distribution Transform (NDT) algorithms (\cite{Magnusson2009}), but all results presented in this publication specifically leverage NDT.  This method excels in environments with large structures, such as buildings and trees, and dynamic terrain.
Figure \ref{fig:orienteer} shows a snapshot of an early match, where the original point-cloud is shown in red, and the aligned point-cloud is shown in blue. 
The magnitude of this correction is due to initialization inaccuracies.
As the pose estimate converges, the corrections become much smaller.

Accurately estimating the quality of the point-cloud registration is necessary to generate a covariance for effectively including pose estimates into the global filter. This has been a challenge with the NDT algorithm. Multiple methods for confidence estimation have been proposed. \cite{Magnusson2009} explored NDT score ($Q_{S}$), largest value of the inverse Hessian ($Q_{H}$), and mean squared point-to-point distance ($Q_{E}$) as potential measures, finding $Q_{S}$ and $Q_{H}$ to have the best performance.
\cite{Wen2018} took the approach of heuristically combining multiple metrics, $Q_{E}$, the convergence time ($Q_{T}$),  and the number of iterations ($Q_{N}$). 
\cite{Akai2017} leverage the inverse Hessian to define the shape of the covariance and scale this distribution by a noise level estimate based on the ratio of the residual sum and the Mahalanobis sum.

The approach implemented in Orienteer similarly uses the inverse Hessian to define the shape of the covariance, but scaling is performed heuristically by combining multiple methods explored by \cite{Magnusson2009}.
First, $Q_s$ is used to obtain a scaled estimate of the relative quality of this pose estimate in comparison to scores achieved for this environment.
This is accomplished by computing a reference score $\overline{Q_{s}}$ based on the last $k$ matches using a windowed moving average filter and then scaling the current score using a logistic function 
\begin{equation}
    A\ =\frac{1}{\left( 1+\exp\left( -a\left( Q_{s} -   \overline{Q_{s}}\right)\right)\right)}
\end{equation}
The quality of the pose estimate is additionally estimated using $B\ =\ b Q_{H}^{2}$ where the square is used to increase the contribution of large values.
Similarly, the components of the eigenvalue matrix of the inverse hessian are squared $C_{i} \ =\ \lambda _{i}^{2}$.
Finally, the diagonal components of the covariance matrix are estimated as a combination of these factors $\sigma _{i} =C_{i}/{(AB)}$.

\section{Experimental Results} \label{sec:experimental_results}
The localization accuracy of TRAELS is computed across a dataset consisting of multiple episodes from 4 different locations, spanning a range of environments, as shown in Table \ref{tab:dataset}.
Filtered state estimates are provided from the GQ7 RTK GNSS-INS which are then fused with wheel odometry using an EKF to produce the ground-truth state estimate.
The covariances of the wheel odometry used in the ground-truth filter are inflated, as described in Section \ref{sec:wheelslip}, and stabilize the pose estimate caused by intermittent GPS outages and jitter in the INS provided estimate.
With an unobstructed view of the sky, the ground-truth achieves an estimated sub \SI{10}{\centi\meter} accuracy, however, in some cases accuracy is degraded e.g. when traveling under dense tree cover.
This does not account for any possible errors related to discrepancies between GNSS and the aligned \textit{a priori} data.

For all Desert episodes (A series), RTK is not available, and the GEO-FOG device with GNSS enabled is used in place of the GQ7 as relatively better, although degraded,  ground-truth performance is observed.
As GNSS was enabled on the GEO-FOG device for these episodes, it is possible that the heading estimate produced by the on-device filter may be influenced by GNSS in addition to gyro-compassing.
However, inspection of the data reveals significant heading error leading us to believe that the heading is solely derived from gyro-compassing.
It is hard to conclude this with certainty due to the black-box nature of the device and a lack of recorded device metadata.
Separately, in all Urban episodes (D series), due to a malfunctioning antenna, only one GNSS antenna is available to the GQ7, which may slightly reduce the accuracy of the ground-truth.

\begin{table}[p]
    \centering
    \begin{tabular}{m{0.03\textwidth} m{0.3\textwidth} m{0.19\textwidth} m{0.19\textwidth} m{0.19\textwidth}}
        \toprule
        \bfseries Env. & \multicolumn{1}{c}{\bfseries Episode Trajectories} & \multicolumn{2}{c}{\bfseries Scene} & \bfseries Description \\
        \midrule
        
        \rotatebox{90}{Desert} &
        {\includegraphics[height=5.35cm]{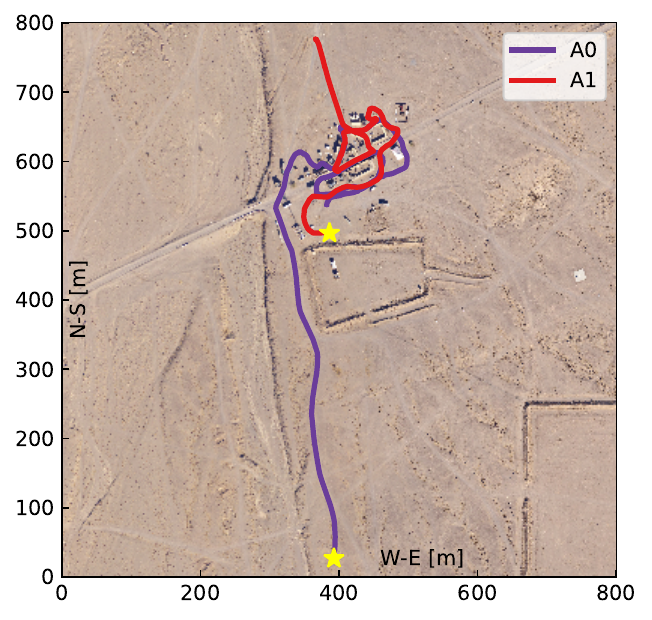}} &
        {\includegraphics[height=5.35cm]{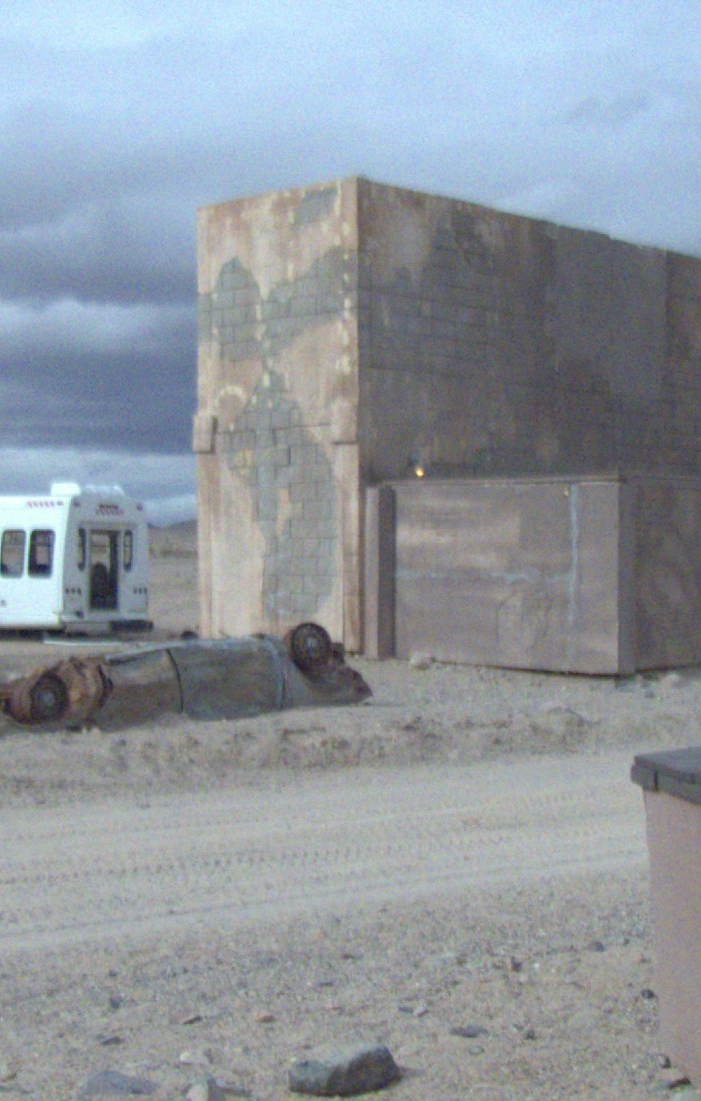}} &
        {\includegraphics[height=5.35cm]{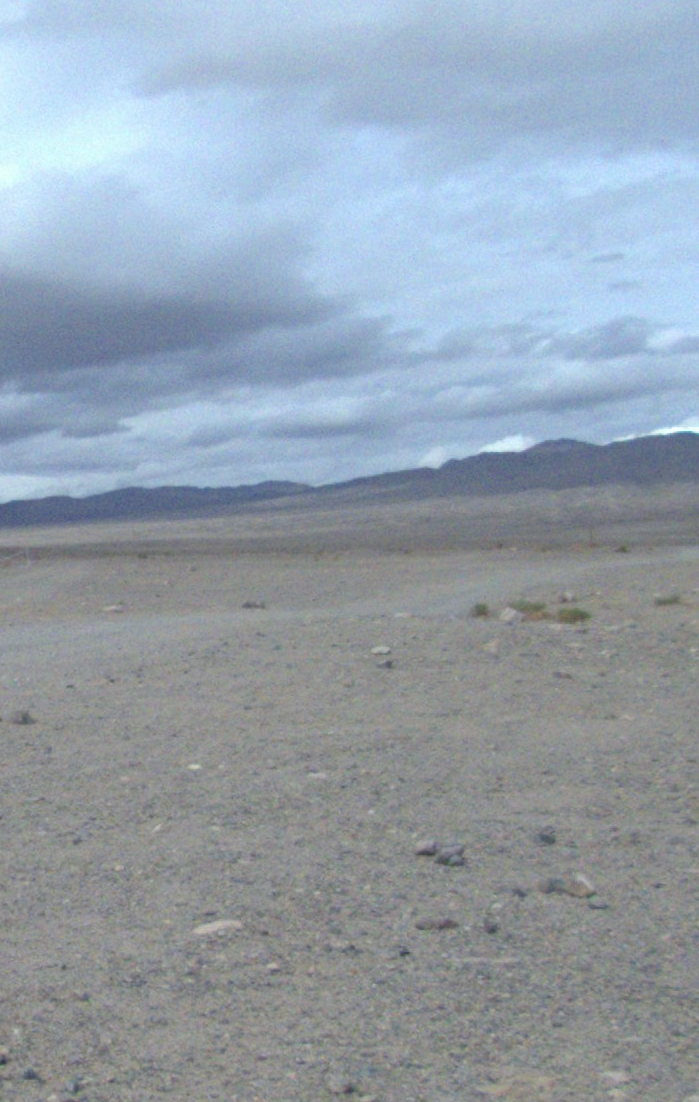}} &
        {Mojave Desert, October: Large flat area with a deserted village. Loose, rocky terrain, sparse vegetation. Frequent small ruts and washouts, creating navigation challenges. Recent re-grading and weather generated significant discrepancies from \textit{a priori} data.}\\
        \midrule
    
        \rotatebox{90}{Forest} &
        {\includegraphics[height=5.35cm]{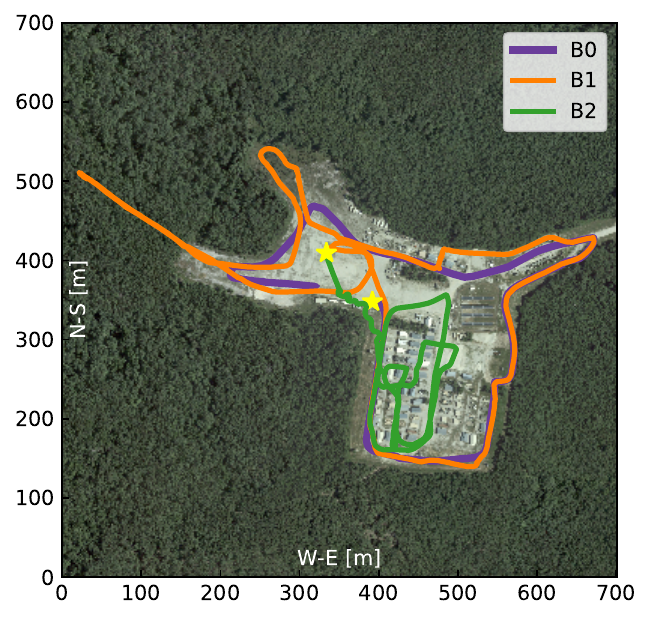}} &
        {\includegraphics[height=5.35cm]{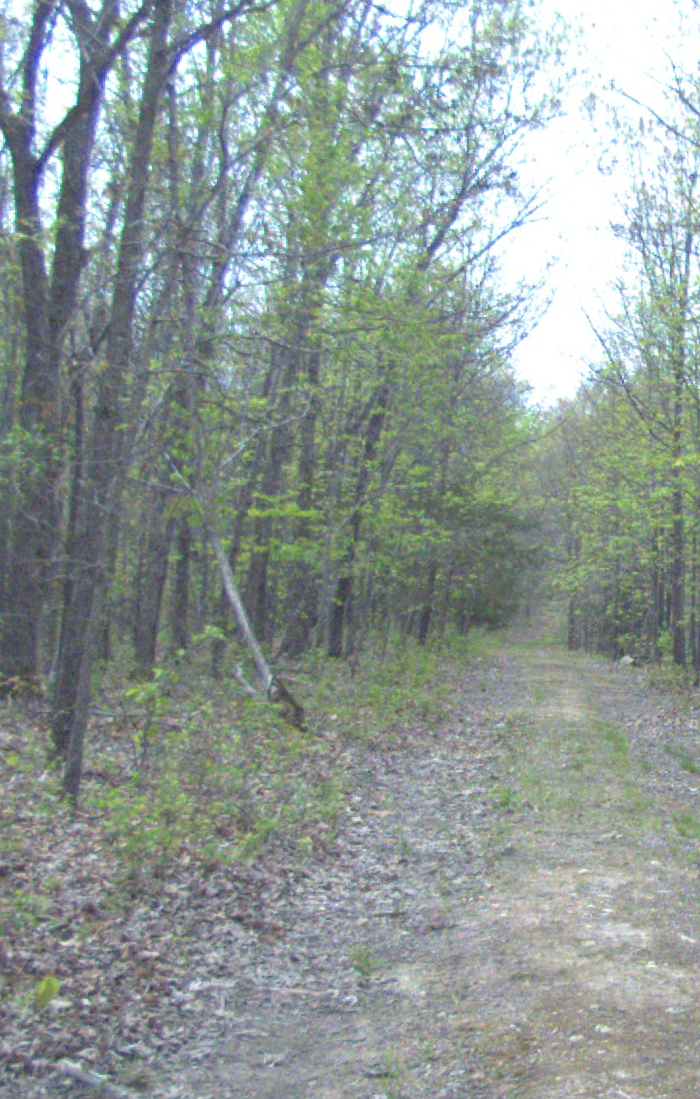}} &
        {\includegraphics[height=5.35cm]{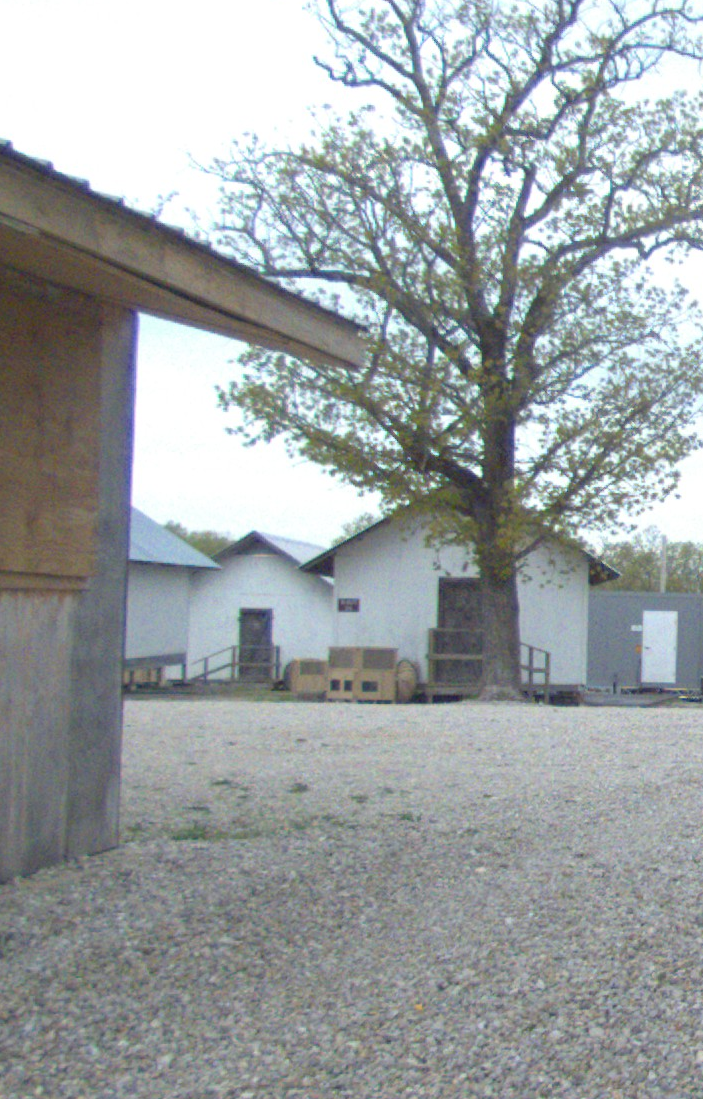}}&
        {Central Missouri, May: Wooded clearing with a collection of buildings. Gravel and unmaintained dirt roads. Recent construction generated significant discrepancies from \textit{a priori} data.}\\
    
        \midrule
    
        \rotatebox{90}{Lake} &
        {\includegraphics[height=5.35cm]{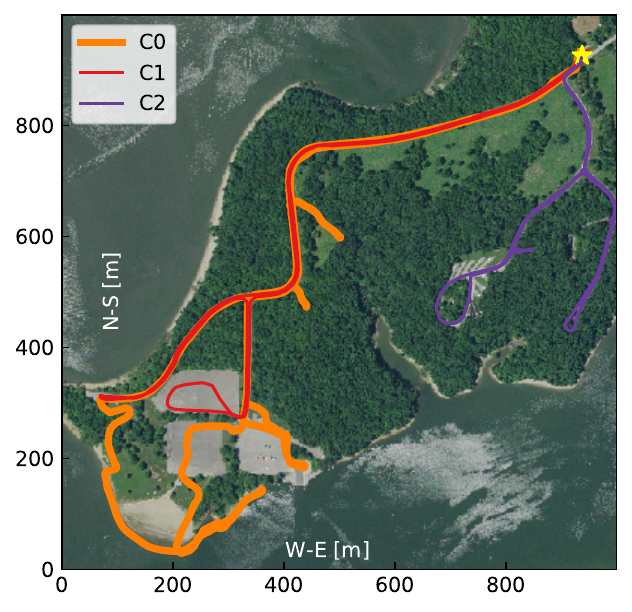}} &
        {\includegraphics[height=5.35cm]{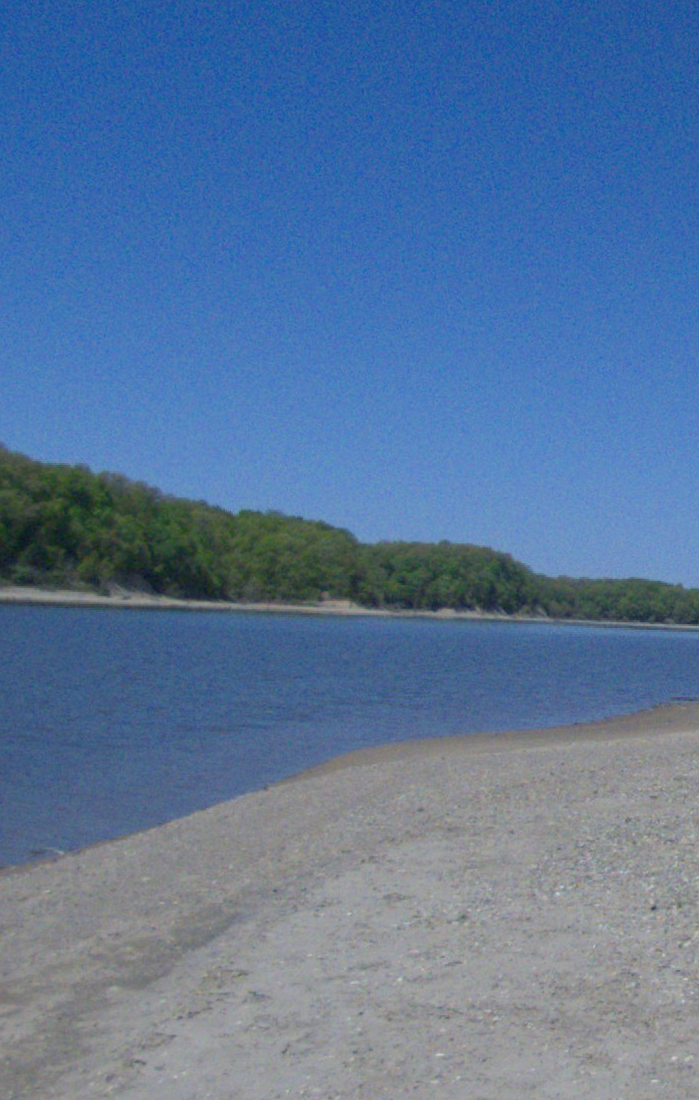}} &
        {\includegraphics[height=5.35cm]{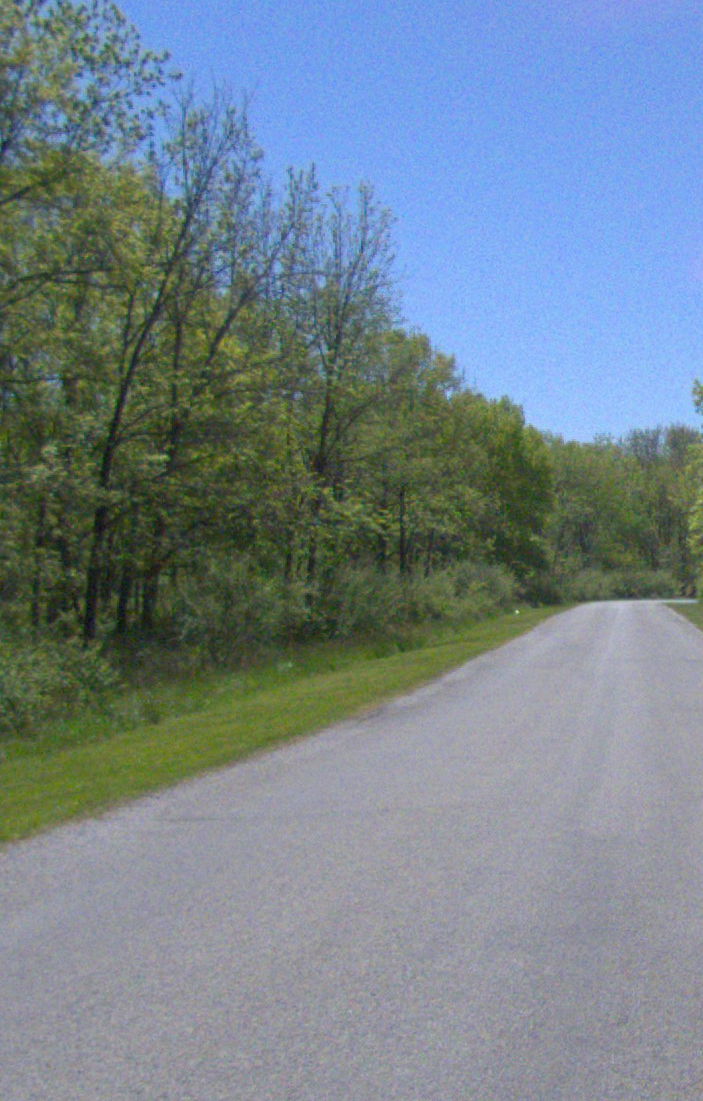}}&
        {Rural Southern Illinois, May: Wooded Peninsula and Lake. Paved roads, off-road, wooded trails, beach. Varying water levels create discrepancies from \textit{a priori} data near the shore.} \\
    
        \midrule
    
        \rotatebox{90}{Urban} &
        {\includegraphics[height=5.5cm]{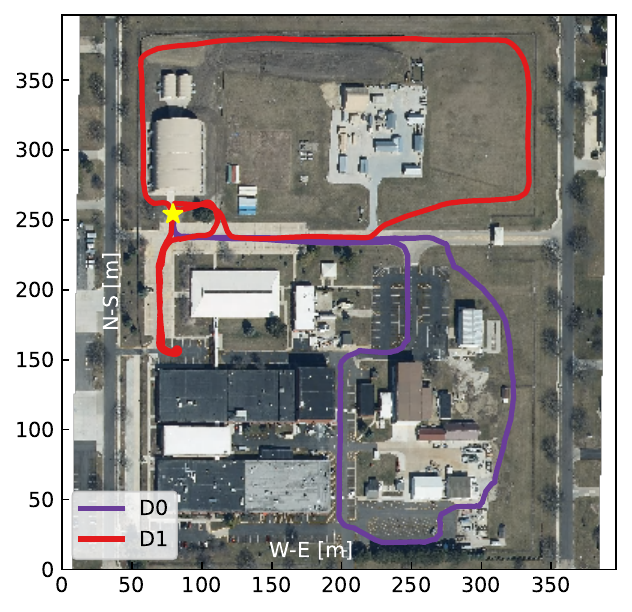}} &
        {\includegraphics[height=5.5cm]{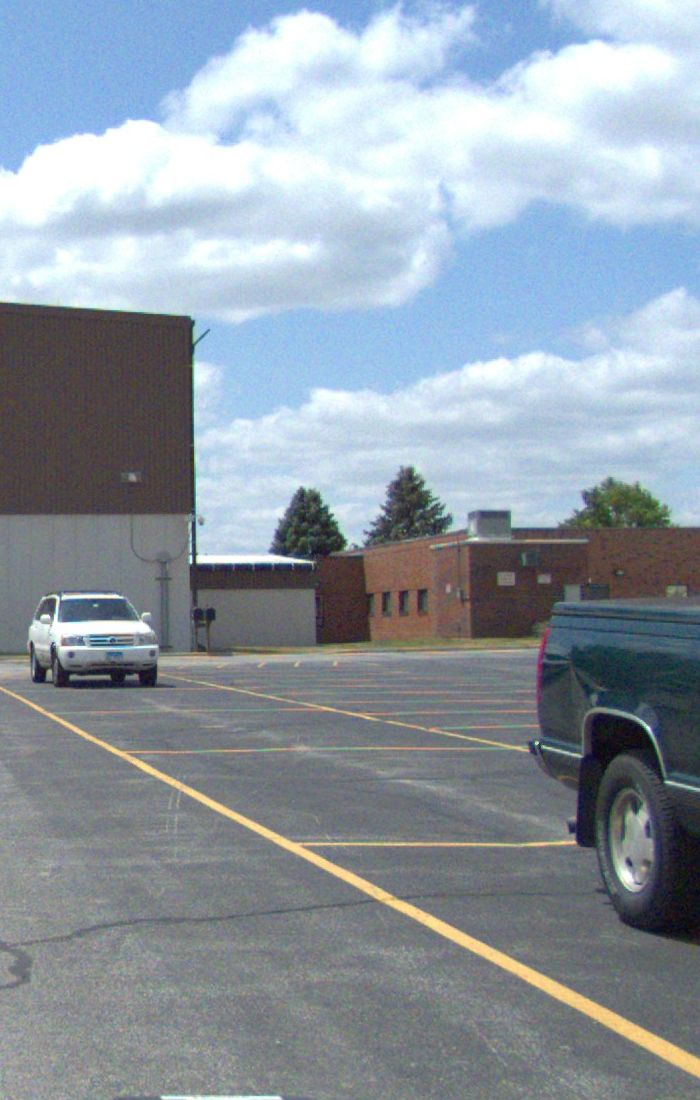}} &
        {\includegraphics[height=5.5cm]{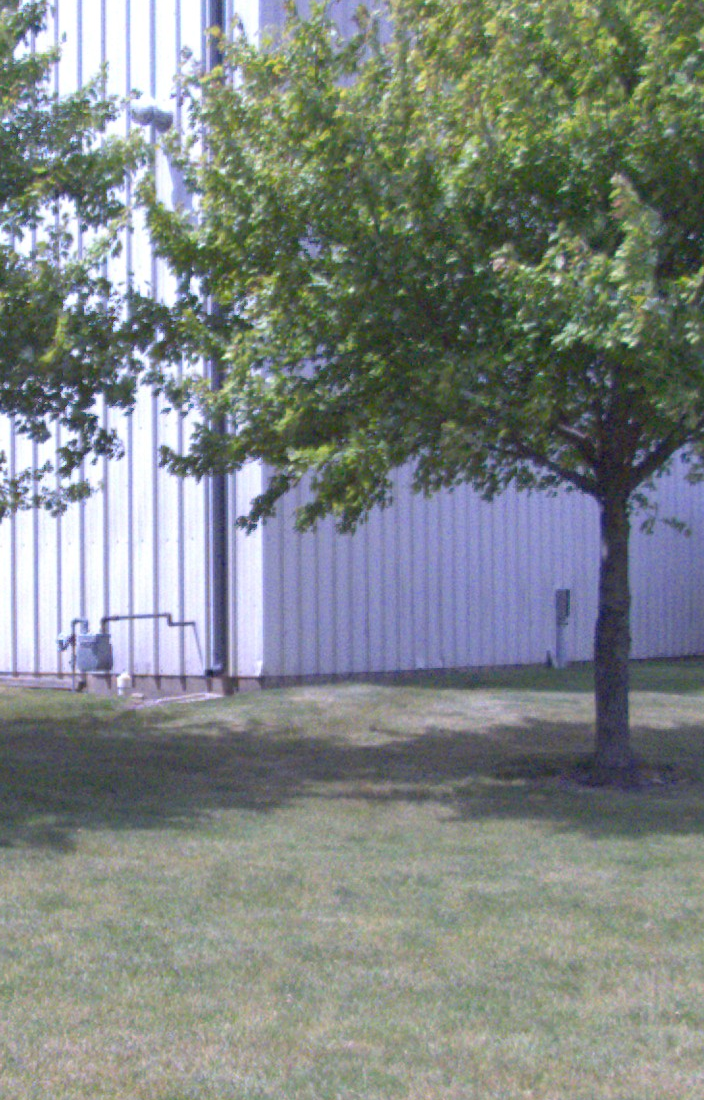}}&
        {Central Illinois, June: distinct buildings and trees, paved roads. Recent construction generated significant discrepancies from \textit{a priori} data.}\\
    
        \bottomrule
    \end {tabular}
    \caption{Overview of dataset organized by environment: Ground-truth trajectories for each episode are visualized with start locations indicated with \textcolor{yellow}{$\bigstar$}, two representative images from each environment are shown, and a description of the terrain is provided.}
    \label{tab:dataset}
\end{table}

\subsection{Metrics}
Two primary metrics are used to quantify the system performance: the absolute trajectory error (ATE) and relative pose error (RPE) \citep{Sturm2012, Meyer2021}.
The ATE is a global metric that captures positional error at each point in time, while the RPE is a local metric describing the relative accuracy of the localization system in a small region over time. Let $\boldsymbol{p}_{t_k} $ and $\boldsymbol{\hat{p}}_{t_k}$ refer to the ground-truth pose and estimated pose at time $t_k$ respectively. Additionally, let $d_{t_k}$ and $\hat{d}_{t_k}$ be the distance traveled from the starting pose according to the ground-truth and the estimator respectively. The metrics are defined as follows
\begin{align}
    \text{ATE}_k &= ||{\boldsymbol{p}_{t_k}} - \boldsymbol{\hat{p}}_{t_k}||
    \label{eq:ATE step}\\
    \text{RPE}_k &= \frac{\Delta \hat{d}_{k} - \Delta{d}_{k}} {\Delta {d}_{k}} \cdot 100
    \label{eq:RPE step}
\end{align}
where $\ATE_k$ has units of meters, $\RPE_k$ is expressed as a percent drift, and $\Delta {d}_{k} ={d}_{t_k}-{d}_{t_{k-1}}$ is the difference in distance traveled between timesteps $t_{k-1}$ and $t_k$.
The distance traveled is computed by integrating the position $d_t = \sum_{k \in [0,t_k]} ||{\boldsymbol{p}_{t_k}} - \boldsymbol{p}_{t_{k-1}}||$, where the filter produces a pose at an average rate of $\SI{10}{hertz}$.
Integrating the distances at this rate instead of the the metric sampling rate determined by $d_{\text{min}}$ enables more accurate comparison of distances traveled for curved trajectories.
However, if a filter produces noisy pose estimates this can lead to overestimation of the distance traveled comparatively.

In this work, the performance of TRAELS is evaluated in terms of its online absolute positional accuracy and its odometric drift. Therefore, the implementation of ATE and RPE metrics differ from the ones used by \citet{Sturm2012} and \citet{Meyer2021}, which are more suitable for assessing the relative/local accuracy of the system when the vehicle is traveling at a fixed speed.
Firstly, no retroactive alignment between the estimated trajectory and the ground truth trajectory is performed prior to computing the ATE, as the goal is to develop globally accurate maps online, not just locally accurate maps. For consistency, in all experiments performed, the position UGV in the global filter is initialized using the RTK GPS and the initial heading is set using the gyro-compassing system after performing the north-seeking calibration procedure. In environments where GPS is truly denied, an operator selects an initial position on a satellite map interface to set the latitude and longitude. The altitude is then obtained by acquiring the elevation from the \emph{a priori} elevation data at that location. 

Secondly, as opposed to evaluating the metrics at a fixed time interval $\Delta t$, errors every $\Delta d_{\text{min}} = \SI{1.0}{\meter}$ along the ground-truth trajectory are analyzed.
Using a fixed time interval provides a nice interpretation of the RPE as a drift velocity if RPE is normalized by $\Delta t$.
However, this also makes RPE and ATE velocity dependent as higher velocities yield larger distances traveled. 
For the step-wise metrics, $\RPE_k$ and $\ATE_k$, distance based selection for non-constant velocity episodes simplifies interpretation of the metrics and helps to reduce the effect of noise on the results. For example, if low amplitude positional noise exists in either the truth or estimated localization estimates, then when traveling at low-speeds, the RPE will be very noisy with the fixed timestep method. When computed over \SI{1.0}{\meter} instead, the effect of this noise on RPE is reduced. 
For the aggregate metrics, e.g. mean and median RPE and ATE, there is a dependence on the velocity traveled during the episode as relatively more measurements are obtained at lower speeds than at higher speeds. This causes aggregate metrics to be biased to be smaller when traveling at lower speeds even when there is no underlying change in the accuracy of the approach at different speeds.
Minimum distance based sampling alleviates this problem. 
This is particularly important for the evaluated dataset as the UGV is often stationary for minutes at a time throughout the episodes, typical for the applications motivating this work.

In all experiments, analysis is restricted to 2D, $\boldsymbol{l}_{xy}$, as the vertical error is normally small with properly registered \textit{a priori} elevation data; the exception being cases of significant horizontal construction or intermittent GNSS outages, which can particularly affect altitude estimates, confounding the analysis.
For run B0, the vertical ATE was computed with a mean of 0.46 and standard deviation of 0.25; this is consistent with the expected accuracy given the \SI{0.3}{\meter} resolution of the \textit{a priori} data.

\subsection{Results}
\begin{figure}[H]
    \centering
    \includegraphics[width=3.1in]{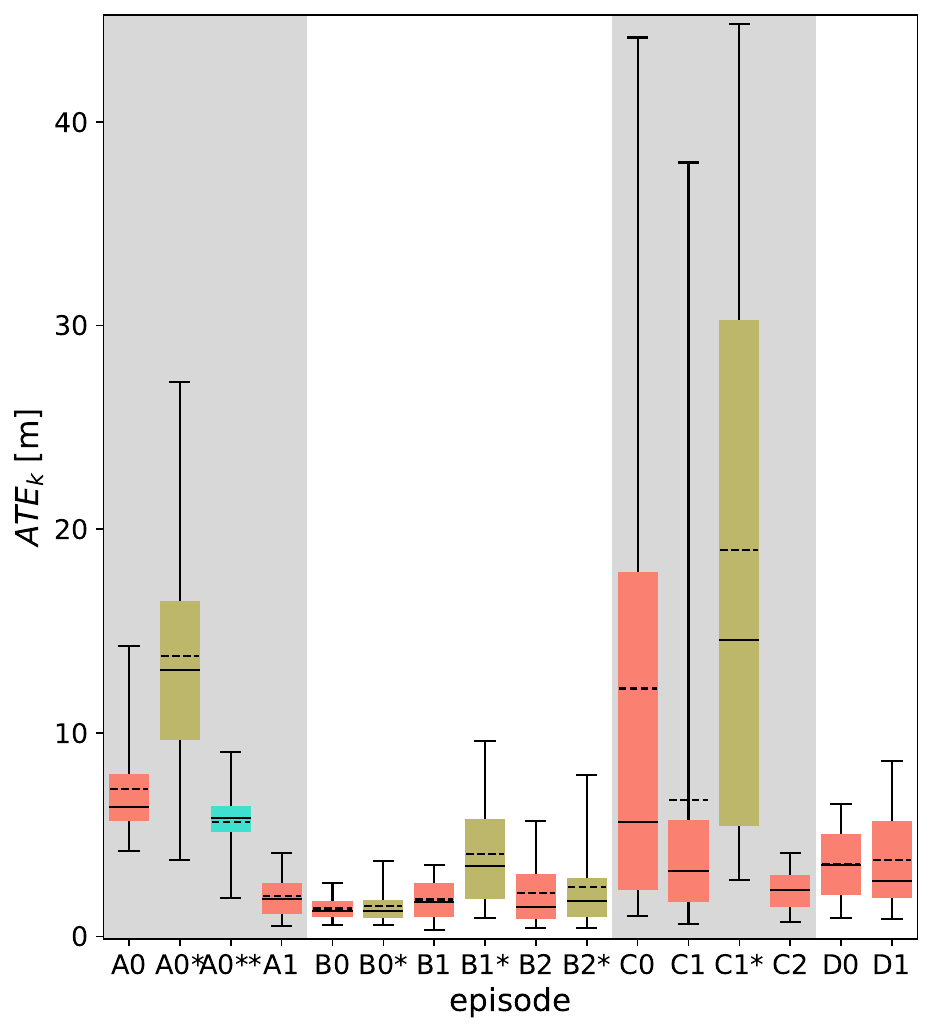}
    \includegraphics[width=3.2in]{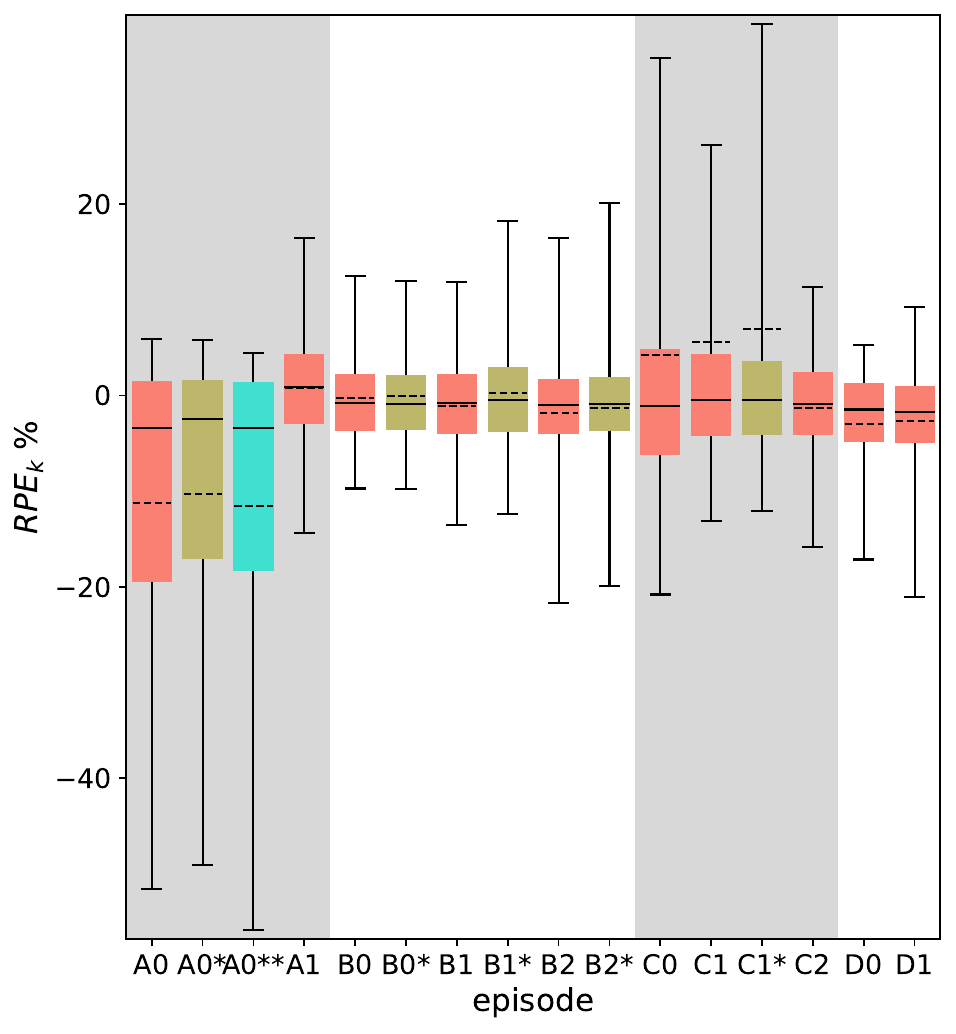}
    
    \caption{Distribution of (left) $\ATE_k$ and (right) $\RPE_k$ for each episode. Boxes extend from the \nth{1} to \nth{3} quartile with whiskers displaying the \nth{5} and \nth{95} percentile. Data outside this range is not shown. Solid horizontal lines indicate the median and dashed the mean. Red boxes are for nominal configuration of TRAELS, green indicates YBE disabled, and cyan indicates fixed $\Delta \hat{\psi}$.}
    \label{fig:result_boxplot}
\end{figure}

The quantitative performance of TRAELS is presented in Table \ref{tab:localization_results} and Figure  \ref{fig:result_boxplot}.
Overall, the results show robust GNSS-denied localization of the UGV across a range of challenging environments.
Where some combination of distinctive geometric

\newpage
\definecolor{Gray}{gray}{0.9}
\begin{table}
    \centering
    \begin{tabular}{|p{0.08\textwidth}|p{0.08\textwidth}|p{0.11\textwidth}|p{0.11\textwidth}|p{0.11\textwidth}|p{0.11\textwidth}|p{0.11\textwidth}|p{0.11\textwidth}|}
        \hline 
         trajectory &  
         length [m] & 
         median \newline velocity [m/s] &
         median \newline \( |{{\RPE}_k}| \% \) &
         max \newline \( |{{\RPE}_k}| \% \) & 
         median \newline \( {{\ATE}_k} \) [m] & 
         max \newline \( {{\ATE}_k} \) [m] & 
         final  \newline \( {{\ATE}_k} \) [m] \\
        \hline 
        \rowcolor{Gray} 
        A0\textsuperscript{\textdagger}	& 982 & 1.0 & 6.2 & 153.2 & 6.37 & 15.21 & 6.32 \\
        \rowcolor{Gray}
        A0*\textsuperscript{\textdagger}	 & 982 & 1.0 & 5.4 & 541.3 & 13.08 & 28.37 & 8.11 \\
        \rowcolor{Gray}
        A0**\textsuperscript{\textdagger}	 & 982 & 1.0 & 5.3 & 99.9 & 5.81 & 9.80 & 6.03 \\
        \rowcolor{Gray}
        A1\textsuperscript{\textdagger}		& 1093 & 1.3 & 3.7 & 117.1 & 1.84 & 5.97 & 5.97 \\
        B0 & 763 & 0.6 & 2.9 & 100.0 & 1.25 & 3.39 & 1.16 \\
        B0* & 763 & 0.6 & 2.9 & 88.6 & 1.24 & 4.56 & 1.21 \\
        B1	& 877 & 0.8 & 3.1 & 144.2 & 1.69 & 9.79 & 1.44 \\
        B1*	& 877 & 0.8 & 3.3 & 224.8 & 3.46 & 11.34 & 3.60 \\
        B2	& 877 & 1.1 & 2.8 & 117.3 & 1.45 & 8.27 & 3.70 \\
        B2*	& 877 & 1.1 & 2.9 & 100.0 & 1.74 & 9.21 & 7.90 \\
        \rowcolor{Gray}
        C0 & 1890 & 1.3 & 5.5 & 1081.0 & 5.61 & 56.84 & 50.35 \\
        \rowcolor{Gray}
        C1	& 1364 & 1.0 & 4.1 & 2729.2 & 3.23 & 55.71 & 0.68 \\
        \rowcolor{Gray}
        C1*& 1364 & 1.0 & 3.8 & 2359.6 & 14.55 & 50.31 & 45.25 \\
        \rowcolor{Gray}
        C2	& 1318 & 1.1 & 3.4 & 285.0 & 2.26 & 6.51 & 0.68 \\
        D0\textsuperscript{\textdagger}	 & 1094 & 1.3 & 3.0 & 99.8 & 3.52 & 6.86 & 3.38 \\
        D1\textsuperscript{\textdagger}	 & 1656 & 1.3 & 3.3 & 88.2 & 2.73 & 11.29 & 2.39 \\
         \hline
    \end{tabular}
    \caption{TRAELS performance across the dataset. Episodes with * indicate that YBE is disabled and ** indicates a fixed $\Delta \hat{\psi}$. \textdagger\ denotes degraded ground-truth performance as discussed in Section \ref{sec:experimental_results}. Ground-truth estimates are noisier for all Desert (A) episodes due to use of a different GNSS-INS resulting in a misleadingly high RPE metric. The noise has less of an impact on the $\RPE_k$ when a point-to-point method as opposed to an integrating method is used for computing the metric, e.g. reducing episode A0 median $|RPE_k|$ from \SI{6.2}{\percent} to \SI{3.1}{\percent}.}
    \label{tab:localization_results}
        
\end{table}

and ground surface features are present, the developed TRN methods are leveraged by TRAELS consistently achieve an $\ATE_k < \SI{3.0}{\meter}$ as demonstrated by the performance in a subset of the episodes, but particularly the Forest environment.
From the low median $\RPE_k$, it is also evident that the dead-reckoning of the system performs well, enabling less frequent TRN corrections to maintain accurate positioning.
When corrections are received, jumps in position sometimes occur resulting in temporarily large $\RPE_k$ as evident by the high values in the max $|\RPE_k|$ column in Table \ref{tab:localization_results}.

The most accurate performance, lowest median $\ATE_k$, is achieved in episode B0 at the Forest location where $\ATE_k = \SI{1.25}{\meter}$ and $\RPE_k = \SI{2.9}{\percent}$.
This environment is challenging for terrain referenced methods due to the presence of repetitive structures and significant environment changes resulting from construction after the time \textit{a priori} data was collected.
The least accurate performance is observed in episode C0, where the final $\ATE_k = \SI{50.35}{\meter}$, reflecting the failure of the system to maintain the state within a range easily correctable using the TRN methods.
TRAELS struggles throughout this episode for a number of reasons.
For a large portion of the episode, the UGV is traveling on a straight paved road through a densely forested area.

ATLIS is able to keep the UGV localized on the road for the majority of the time, but pulls the estimate off longitudinally (along the direction of travel), indicating an underestimation in the uncertainty of the ATLIS pose estimate.
Orienteer similarly struggles in these regions, due to ambiguity in the geometric information afforded by the environment.
Midway through the episode, the UGV travels along a beach before returning back along the same path.
Due to changes in the lake water level from the time the \textit{a priori} data was obtained, the shoreline is at a significantly different location, estimated to be on the order of \SI{10}{\meter}, which causes ATLIS to erroneously pull the vehicle toward the edge of the beach in the \textit{a priori} image.
When present in a single episode, these environmental characteristics prove to be more than the system can handle.
However, in episode C2 when the route is shortened and the UGV does not approach the shoreline, the system achieves much better performance and demonstrates the ability of the system to recover from very large errors.

\begin{figure}
    \centering
    \includegraphics[width=3.8in]{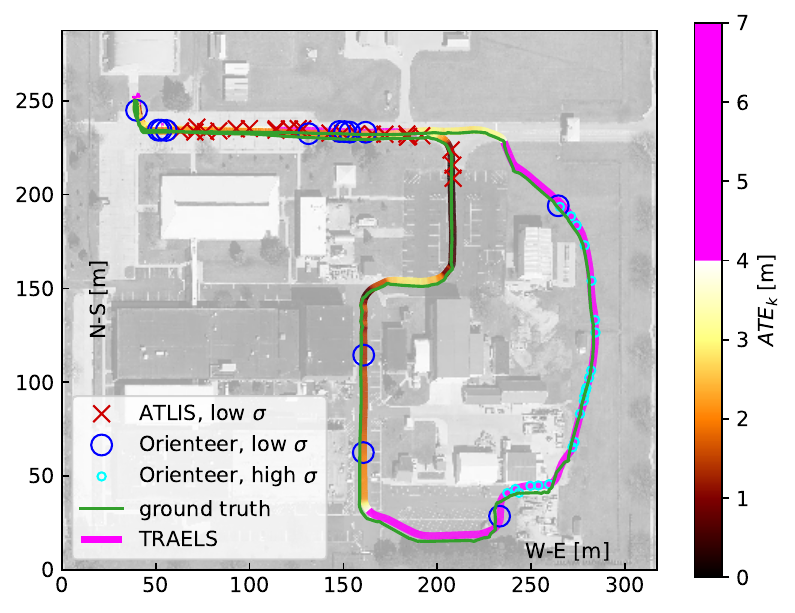}
    \includegraphics[width=2.75in]{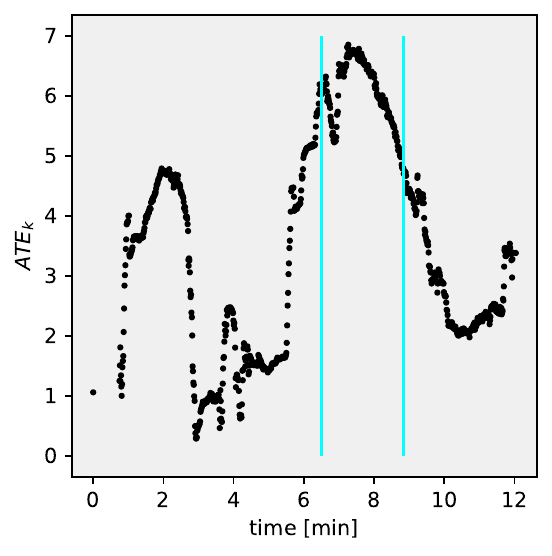}
    \caption{TRAELS performance for run D0 where (left) ATLIS and Orienteer corrections with low $\sigma_x^2 \Vert \sigma_y^2 < \SI{10}{\meter\squared}$ covariance in either geographical direction (N-S, W-E) are shown. For illustration purposes, ATLIS corrections are decimated. Low confidence Orienteer fixes are additionally shown in the region in the east of the map beginning and ending with high confidence Orienteer fixes, (right) also marked with vertical turquoise lines on the $\ATE_k$ plot.}
    \label{fig:urban_atlis_orienteer_overaly} %B2 overlay
\end{figure}

Figure \ref{fig:urban_atlis_orienteer_overaly} helps to illustrate how ATLIS and Orienteer can be complementary.
In the areas where the UGV is close to buildings, Orienteer finds good matches with the \textit{a priori} map because the abundance of vertical features constrain the registration well.
In the regions where there is significant variation in ground color, ATLIS provides low covariance estimates, and in the regions where the ground surface is more uniform, it estimates higher uncertainties.
The $ATE_k$ rises sharply at the bottom of the map around the \SI{6}{\minute} mark.
At this location, the vehicle is performing a gradual turn, and the wheel covariances are increased according to Equation \ref{eq:wheel_cov} allowing the accelerometer measurements to have a greater influence on the estimation.
However, because the EKF does not perform lever arm compensation, the centripetal acceleration may cause the position estimate to drift temporarily.
In the East region of the map in the grass with few nearby structures, no ATLIS corrections are observed and only high covariance estimates are produced by Orienteer.
The $ATE_k$ begins to decrease approximately at the midpoint of the highlighted region.
This may be explained by a heading bias or wheel diameter error which would cause the error to decrease as the UGV gets closer to the starting location. 
When the UGV rounds the North side of a building, the point-cloud is more geometrically rich enabling Orienteer to produce a more confident pose estimate.
Also note that in this region, the estimated trajectory is only displaced a short distance laterally from the ground-truth, but the $\ATE_k$ is high, indicating significant error along the direction of travel of the vehicle.
This is interesting because it allows us to conclude that the high covariance pose estimates that Orienteer is producing here are still enough to keep the lateral distance between the buildings and the vehicle correct, but not informative enough to correct the longitudinal error.

\begin{figure}
    \centering
    \includegraphics[width=3in]{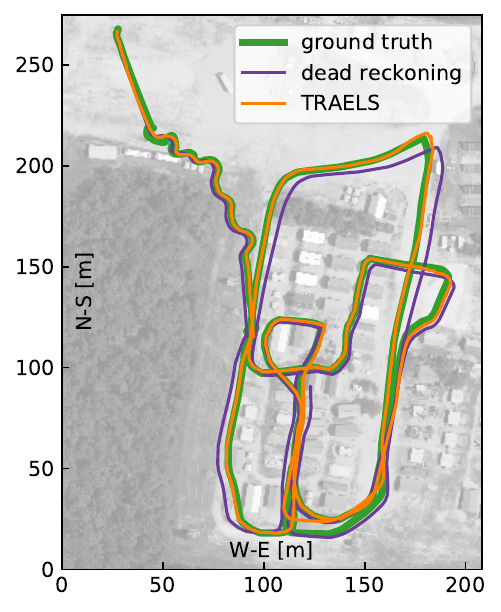}
    \includegraphics[width=3.71in]{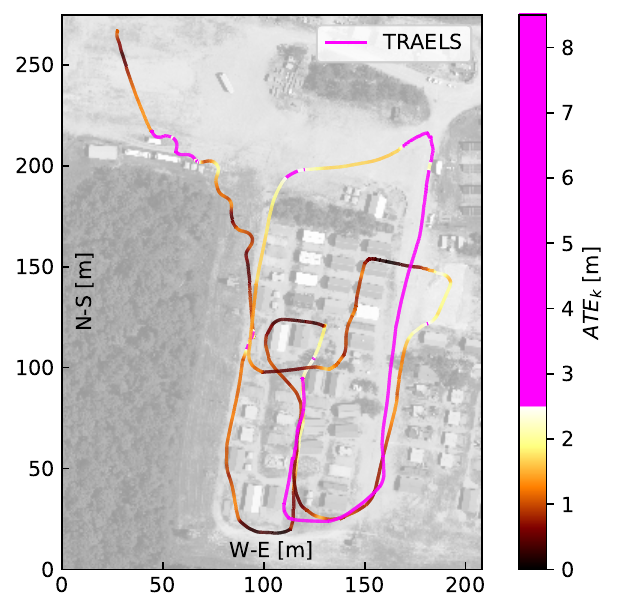}
    
    \caption{TRAELS performance for episode B2. (left) Overlay of trajectories where (right) $\ATE_k$ is used to color the trajectory.}
    \label{fig:B2 ATE}
\end{figure}

In episode B2, the $ATE_k$ is relatively low when traveling through the cluster of buildings, as shown in Figure \ref{fig:B2 ATE}.
As the vehicle travels into the open area in the North-East corner of the map, the error grows until it re-enters the area with the buildings again where the TRN methods can correct for errors.
Similar to episode D0, high longitudinal and low lateral error is observed along the east edge of the map as indicated by the TRAELS trajectory overlapping the ground truth and $\ATE_k$ being large.
A quick jump in $\ATE_k$ is also observed when the UGV makes a sharp right turn and is likely a negative effect of the wheel slip rejection.

\begin{figure}
    \centering
    \includegraphics[width=2.5in]{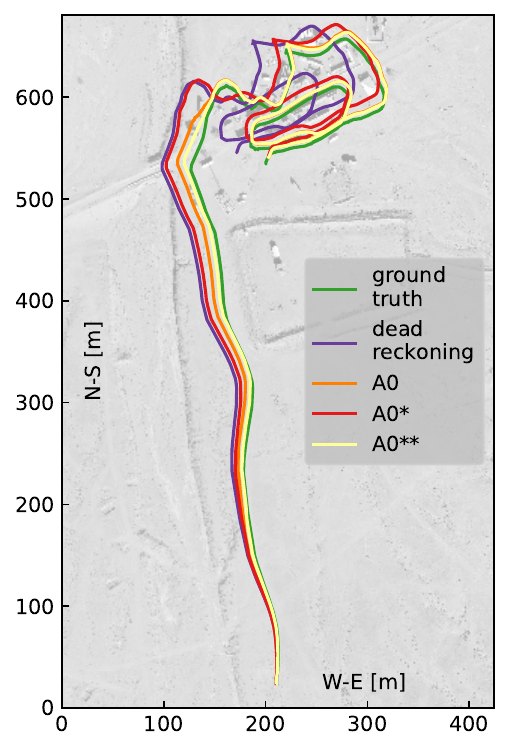}
    \includegraphics[width=3.6in]{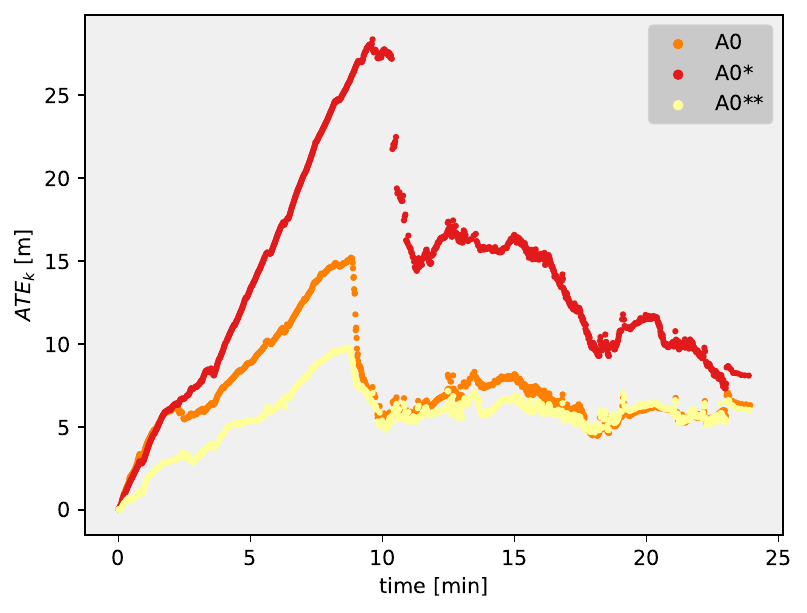}
    
    \caption{YBE ablation test for episode A0. (left) Trajectory overlay and (right) performance indicated by $\ATE_k$. Even when starting in an area not conducive for confident terrain-referenced corrections, YBE is able to dampen drift caused by heading error. After arriving in a feature-rich region, system performance is comparable to A0**.}
    \label{fig:YBE}
\end{figure}

To examine the effect of yaw bias $\Delta \psi$ on the overall system performance, an ablation was performed for a subset of the episodes (A0, B0, B1, B2, C1) where YBE was disabled as indicated by the * in Table \ref{tab:localization_results}.
As can be seen in Figure \ref{fig:YBE}, when traversing long distances across areas lacking distinctive terrain features, a bias in initial UGV heading can result in large translational errors.
With the use of YBE, even the sparse low confidence corrections received from ATLIS matching to a barely visible trail, yield a difference in local and global trajectories and subsequently improve estimation of $\Delta \hat{\psi}$, reducing the $\ATE$ substantially.
By the point the UGV reaches a more feature rich portion of the environment, the TRN methods are able to more quickly correct for the accumulated drift than when YBE is not employed.
With YBE running, approximate convergence of $\Delta \hat{\psi}$ was observed on most episodes.
Therefore, another experiment was performed (A0**) where the yaw bias correction was fixed to be the final yaw bias estimate from run A0 $\Delta \hat{\psi}_{A0} = \SI{1.37}{\degree}$.
In this configuration, the $\ATE$ is reduced even further than when YBE is employed.
The error still grows at a relatively fixed rate until reaching the buildings indicating a small error in odometry likely caused by a wheel slip or a wheel diameter calibration inaccuracy.
This analysis reveals the importance of accurately estimating global heading particularly in feature sparse environments. 

Without a known high accuracy ground-truth for orientation, it is difficult to precisely determine the absolute accuracy of the gyro-compassing derived heading estimates. However, for episodes where gyro-compassing is presumed to successfully estimate orientation with high accuracy, or ATLIS and Orienteer are able to regularly make confident matches, YBE does not have a significant impact (B episodes). When substantial yaw biases do exist, and/or there are long periods with few confident TRN matches, (A and C episodes), YBE can lead to a significant reduction in error if at least some TRN corrections are provided.
Additionally, for ATLIS in particular, heading errors can result in a large positional offset.
For example, in areas where features are radially symmetric, such as gently curved roads, the theoretical best match is found at a point translated along the length of the feature's cord producing an error of magnitude $2r\sin(\Delta \psi /2)$ which can be quite large for even small yaw errors.

When starting out in feature-sparse environments, if the uncertainty of the heading measurement is underestimated, the global EKF will produce overconfident position estimates.
Given that the ATLIS search window is coupled to positional uncertainty, this overconfidence can cause to the true position to be out of range and lead to the inability of the ATLIS to counteract this accumulating error. 
Orienteer does not have an explicit search range, however in the tested configuration it is generally unable to provide corrections beyond a few meters, and is more suited at performing more precise alignments. 

\section{Conclusions}\label{sec:conclusion}
In this work, we have detailed and validated an approach to accurate long-range GNSS-denied localization in unstructured environments.
Through careful calibration of inertial sensors and use of dynamic varying wheel odometry covariances wheel-slip rejection, low-drift dead-reckoning is achieved.
By fusing this low-drift odometry with two complementary TRN methods using an EKF, yaw biases can be estimated and robust GNSS-Denied global localization is achieved.
In conducive environments, TRAELS consistently achieves a median $\ATE_k$ of less than 3.0$m$.
Given the uncertain absolute accuracy of the \textit{a priori} maps, this result is compelling.
However, it is possible to achieve superior results through environment specific tuning. For example, by slightly increasing the confidence of Orienteer pose estimates, a mean $\ATE_k = \SI{0.15}{\meter}$ was achieved in a separate test in the Urban environment, which is one-half the resolution of the \emph{a priori} data. Unfortunately, this tuning caused poor results for other environments where Orienteer more frequently produced overconfident matches.

Accurate covariance estimation for all sensors and TRN methods is essential to the achieving good localization performance when using an EKF for localization.
This subsequently leads to accurate estimation of the current state estimate covariance $\sigma_{q_k}$ which is valuable for enabling robust localization and recovery from large accumulated drift in feature-poor environments with TRN methods utilizing dynamic search windows.
Although not analyzed here, the localization performance of the TRAELS system enables the generation of high-quality maps with minimal ghosting.
By dynamically disabling mapping when the pose covariance grows beyond a threshold, generated maps can be ensured to meet certain accuracy requirements.
Large covariances can also be used to trigger a re-initialization procedure to ensure accurate global navigation.
The robustness of the presented approach also enables navigation to distant waypoints.
Because the system is not dependent on loop-closures, accuracy can be maintained over one-way missions without the requirement of returning to previous positions to improve map quality.

Future work is needed to explore alternative filtering methods that more easily allow incorporation of delayed measurements than an EKF, e.g. factor graph approaches, as the TRN methods considered here produce estimates with significant delay. 
New methods should also support inclusion of sensor and mounting biases as part of the filter state to reduce the need for time consuming calibration and simplify the localization architecture.
Ideally, in addition to supporting global pose estimates, localization frameworks should support inclusion of loop-closure constraints, which could help improve performance when \textit{a priori} data is low resolution or outdated.
To improve TRN methods, further work is necessary to better characterize pose estimate uncertainty and reduce erroneous matches that can negatively affect performance.
Use of Monte Carlo Localization methods to pre-filter TRN estimates prior to sensor fusion could enable better enforcement of Gaussian distributional assumptions and more careful consideration of multiple hypotheses and that commonly occur in environments containing symmetry.

\section{Acknowledgments}
The authors thank the Robotics for Engineer Operations team for their support in making this research possible. Dylan Charter and Charles Ellison have been instrumental in ensuring the success of this work. Kenneth Niles and Matthew Richards are primarily responsible for the development of the original ATLIS and SMGS Orienteer packages and participated in many discussions on terrain referenced navigation. Arturo Saucedo helped develop the \textit{a priori} data preprocessing pipeline.

\printbibliography

\end{document}